\documentclass{article}

\usepackage[preprint]{neurips_2024}

\usepackage[utf8]{inputenc} 
\usepackage[T1]{fontenc}    
\usepackage{hyperref}       
\usepackage{url}            
\usepackage{booktabs}       
\usepackage{amsfonts}       
\usepackage{nicefrac}       
\usepackage{microtype}      
\usepackage{xcolor}         
\usepackage{graphicx}
\usepackage{tcolorbox}
\usepackage{makecell}
\usepackage{subfiles}
\usepackage{geometry}
\usepackage{enumitem}
\usepackage[utf8]{inputenc}

\title{Building Understandable Messaging for Policy and Evidence Review (BUMPER) with AI}

\newcommand{\NAMEIT}{BUMPER}
\newcommand{\NAMEITs}{BUMPERs}
\newcommand{\gptthree}{gpt-3.5-turbo-0125}
\newcommand{\gptfouro}{gpt-4o-2024-05-13}
\newcommand{\gptfour}{gpt-4-0125-preview}

\newtcolorbox{fancytextbox}{
    colback=gray!3,
    colframe=black,
    boxrule=0.5pt,
    arc=0pt,
    top=2pt,
    bottom=2pt,
    left=6pt,
    right=6pt
}

\author{%
Katherine A. Rosenfeld\thanks{Institute for Disease Modeling,  Bill \& Melinda Gates Foundation}\;\,\thanks{katherine.rosenfeld@gatesfoundation.org}\\
\And 
Maike Sonnewald \thanks{Department of Computer Science,
University of California, Davis} \\
 \\
\AND
Sonia J. Jindal$^*$ \\
\And 
Kevin A. McCarthy$^*$ \\
\And
Joshua L. Proctor$^*$ \\
}

\begin{document}

\maketitle




\begin{abstract}

We introduce a framework for the use of large language models (LLMs) in Building Understandable Messaging for Policy and Evidence Review (BUMPER). LLMs are proving capable of providing interfaces for understanding and synthesizing large databases of diverse media. This presents an exciting opportunity to supercharge the translation of scientific evidence into policy and action, thereby improving livelihoods around the world. However, these models also pose challenges related to access, trust-worthiness, and accountability.  The BUMPER framework is built atop a scientific knowledge base (e.g., documentation, code, survey data) by the same scientists (e.g., individual contributor, lab, consortium). We focus on a solution that builds trustworthiness through transparency, scope-limiting, explicit-checks, and uncertainty measures. LLMs are rapidly being adopted and consequences are poorly understood. The framework addresses open questions regarding the reliability of LLMs and their use in high-stakes applications. We provide a worked example in health policy for a model designed to inform measles control programs. We argue that this framework can facilitate accessibility of and confidence in scientific evidence for policymakers, drive a focus on policy-relevance and translatability for researchers, and ultimately increase and accelerate the impact of scientific knowledge used for policy decisions.

\end{abstract}

\section{Introduction}
\label{sec:introduction}

A core motivating factor, and source of funding, for many scientific studies is to improve livelihoods in our communities, nations, or world. Budgets for research and development have been steadily growing with the US government alone reaching over 700 billion dollars in 2020 \citep{anderson23}. However, to have impact, a scientific result needs to be communicated effectively \citep{weiss75, weiss79, gluckman16, wilsdon14, elliott21}. Today, the way science is communicated often results in isolated monoliths of evidence, including papers, proceedings, presentations, and datasets, which are detached from the real-world situations and questions that could benefit from their insights. While there are ways for decision makers to request and receive advice from the scientists, they are not straightforward \citep{gluckman21}. We present a framework for the use of large language models (LLMs) in Building Understandable Messaging for Policy and Evidence Review (BUMPER). \NAMEIT\,utilizes a reactive chat interface to bridge the gap between individual scientific studies and their application in, for example, policy. BUMPER is not meant to generate new evidence by, e.g., functional search \citep{romera_parades24} or synthesis \citep{zheng23}; rather, it is a translational tool for existing evidence. It features a novel compliance score and, unlike many popular tools today, establishes clear ownership, enabling accountability and keeping the scientist ``in the loop''. \NAMEIT\, has the potential to empower policymakers to more effectively use evidence and data in their decision-making processes.

Who is at fault if you build a clock and give it to someone seeking a compass, only for them to later wander in circles? The process of transferring science to decision making wrestles with this problem: what mechanisms can ensure that the right information is incorporated into the right application? To address this key issue, \NAMEIT\,is more than just a chatbot. It introduces innovative, scope-limiting and transparent checks that are defined by the scientists, while also ensuring clear attribution and ownership. Unlike many other guardrail architectures \citep{guardrails_ai, inan23, rebedea23, goyal24, dong24}, \NAMEIT\,does not directly affect the output by e.g., filtering or iteration. We argue this pattern promotes trustworthiness and show case studies from a real-world disease modeling analysis and a toy-model.

\begin{figure}[t]
  \centering
  \includegraphics[width=\textwidth]{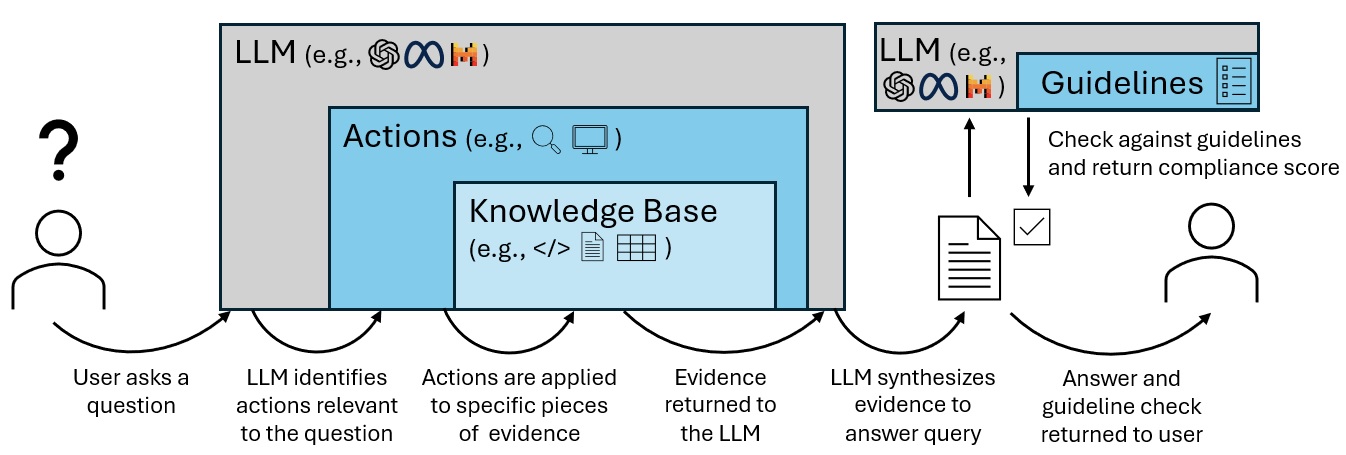}
  \caption{An overview of the \NAMEIT\,framework. See steps 1-5 in section \ref{sec:framework} for details.}
  \label{fig:workflow}
\end{figure}

LLMs promise to transform scientific knowledge transfer by allowing users to pose questions or describe scenarios in familiar language rather than navigating an ever-expanding sea of evidence and technical lingo \cite{lewis21}. However, these models, already in active use, pose challenges related to access, trustworthiness, and accountability \citep{bommasani22}. In particular, LLMs' demonstrated predilection for hallucination is significant concern regarding their use in high-stakes applications \citep{adlakha24,liu22,muhlgay23,li23}. Decision making is one arena where, even the occasional failure, can have severe consequences across the system impacting not only high-impact issues, but also trust and relations between scientists and decision makers. While the problem of LLM hallucinations remains an unsolved challenge, the BUMPER framework offers a promising and novel path forward in the scientific discourse surrounding how we, as a global society, can safely harness the potential LLMs offer. Our contributions include:

\begin{itemize}
    \item Demonstrating how LLMs can be used for scientific evidence synthesis while maintaining direct association and ownership with the scientists
    \item Formalizing a compliance score to assess performance and trustworthiness
    \item Establishing methods to characterize the stability of the framework
    \item Prototyping code and case studies to enable scientists to explore this space
\end{itemize}


In this paper we first describe the \NAMEIT\, framework, highlighting its novel features as well as potential in the space of evidence translation. We then demonstrate the framework first with a simple example of a model for rugby team performance and second a real-world application for health policy with a model of measles seasonality and susceptibility. We conclude with a discussion.


\section{Framework}
\label{sec:framework}
Decision makers have questions, concerns, and pressures. Scientists produce analyses with results, assumptions, and limitations. The role of \NAMEIT\,is to bridge the two via natural language and a well-defined, purposeful architecture. 

The scientist begins with a set of assets composing the \textit{knowledge base}, $K$, and designs a set of \textit{actions}, $A = (a_0, \ldots, a_J)$. Each action, $a_i$, takes a subset of $K$ as inputs along with other optional argument. The outputs of each $a_i$ is information derived from the knowledge base and potentially relevant to the user. Lastly, the scientist must supply guidelines $G$ composed of \textit{criteria}, $C = (c_0,\ldots, c_N)$, and topics, $T = (t_0, \ldots, t_M)$. With these elements, evidence is synthesized in 6 steps: 1) A question from the user is passed to an LLM. 2) The LLM is used to identify which subset of actions, $a_i \in A$, are relevant. 3) The actions are executed and results are aggregated to synthesize an answer using the LLM. 4) Another LLM instance is used test whether the answer complies with the guidelines, $G$ and produces a score. 5) The answer and score are returned to the user. Illustrated in Figure \ref{fig:workflow}, the BUMPER workflow steps 1-5 are:

1) \textit{User interaction:} The user interacts with \NAMEIT\, in a prompt-reply loop or chat format. The text from a user's prompts and the \NAMEIT\,responses are stored in a thread that both the user and LLM have access to (see Figure \ref{fig:measles_example} for an example). Areas of interest a user might have include: methods, scope, risk, comparisons, counterfactuals, situational estimation, financials/cost, and prioritization. However, not all of these topics will be appropriate for any single \NAMEIT\, knowledge base. The role of the guidelines in step 5 is to flag when a synthesized answer lies out of scope.

2) \textit{Action identification:} To create a \NAMEIT\,associated with a scientific analysis, the scientist identifies components of their work that may be relevant to decision makers. For example, this may be particular pieces of documentation, calculations, or simulation results. For each piece, the scientist creates a function whose output is input for the LLM. This is most likely, but not limited to, text and the functions can vary widely from a table lookup to running code \citep{vertex_ai, openai_api}. In addition the functions, the scientist also provides a description of the function describing its purpose and providing context for understanding the result. This meta-data provides the context for matching functions to a query. 

The task of matching actions can be solved in a variety of ways and, under the umbrella term ``agents'' \citep{xi23}, and is an area of rapid development. In our examples we use OpenAI's assistants API which is a black box regarding how the matching is done. However, there do exist open-source methods  combining LLMs with tasks that may extend to this problem \citep[e.g., TALM, PAL, ReAct;][]{parisi22, gao23, yao23, wang23, askitov23}. 


3) \textit{Knowledge retrieval and aggregation}: The results of the selected actions, along with the function meta-data, are aggregated to provide context for answering the original question using an LLM. The resulting outpus is the synthesized evidence, $E$.

4/5) \textit{Evidence scoring and output}: The final step involves generating a flag (pass/fail) and a white-box score, $S$, to indicate whether and how well the answer aligns with the intended purpose of the original knowledge base, $K$. Using an LLM (without the original context), the synthesized evidence is compared against the criteria and topics that make up the guidelines. The criteria list the requirements that must be met (e.g., do not provide financial estimates), while the topics further limit the scope of the answer (e.g., methodology). See case studies for further details with context.

To distinguish between confidence in the correctness of the synthesized answer and confidence in whether the answer is in-scope, we utilise a \textit{compliance score} $S$. To generate $S$, where $S \in [0,1]$, we take the LLM-generated output from the guideline check and de-compose it into its individual tokens (alpha-numeric combinations that concatenate into the output text). For a given input to the LLM, $G$, each returned token has a probability $P_T(G)$ given the previous tokens $T=0,1,\ldots,T-1$. The probability of the first token $P_0$ is a proxy for confidence \citep{guo17, malinin20, jiang21, kadavath22}. We focus on two ways to compute $S$:
\begin{enumerate}
\item A single simultaneous assessment of all the guidelines, $S=P_0$
\item Combining assessments of the guideline components:
\begin{equation}
S = \prod_{i=0}^NP_0(\mathrm{c_i}) \times \left[1 - \prod_{i=0}^M\left(1-P_0(t_i)\right)\right]
\end{equation}

\end{enumerate}

We find that combining assessments of the guideline elements tends to be more consistent over multiple iterations with lower compliance scores. This result is consistent with previous studies showing that while LLMs tend to be overconfident when verbalizing their confidence \citep{xiong24} there exist techniques for addressing this issue. See section \ref{sec:measles} for examples and further discussion. This compliance score \texttt{is not} used to augment the synthesized evidence, but returned alongside it. This is a novel and critical element of \NAMEIT: the score is not meant to improve performance, but to rather to provide a direct conduit for the scientist to remain in the loop and indicate how well the answer conforms with the overarching criteria. Importantly, it is a measure of compliance and not correctness.

Another key option for \NAMEIT\,is to prompt the LLM to output an ``explanation'' of its answer. This provides to the user complementary information that can help de-mystify the pass/fail flag and compliance score. We also see that adding an explanation directive to the LLM prompt impacts the compliance score. Explanation augmented prompts, related to chain-of-thought, have been shown to strong effect a variety of tasks from symbolic reasoning to commonsense question answering \citep{kojima22, lampinen22, jwei23, talmor19}. Understanding the mechanism that causes this phenomenon is the topic of active research, but out of scope for the present paper.

We present \NAMEIT\,as a generalized framework both for the knowledge being transferred, but also note that it can be built from a variety of libraries and languages, the implications of which will be discussed later. Our examples utilize the OpenAI API with \gptfour\, \citep{openai_api}. Code is available on GitHub at https://github.com/krosenfeld/bumper\_paper.

\begin{figure}[t]
  \centering
  \includegraphics[width=0.9\textwidth]{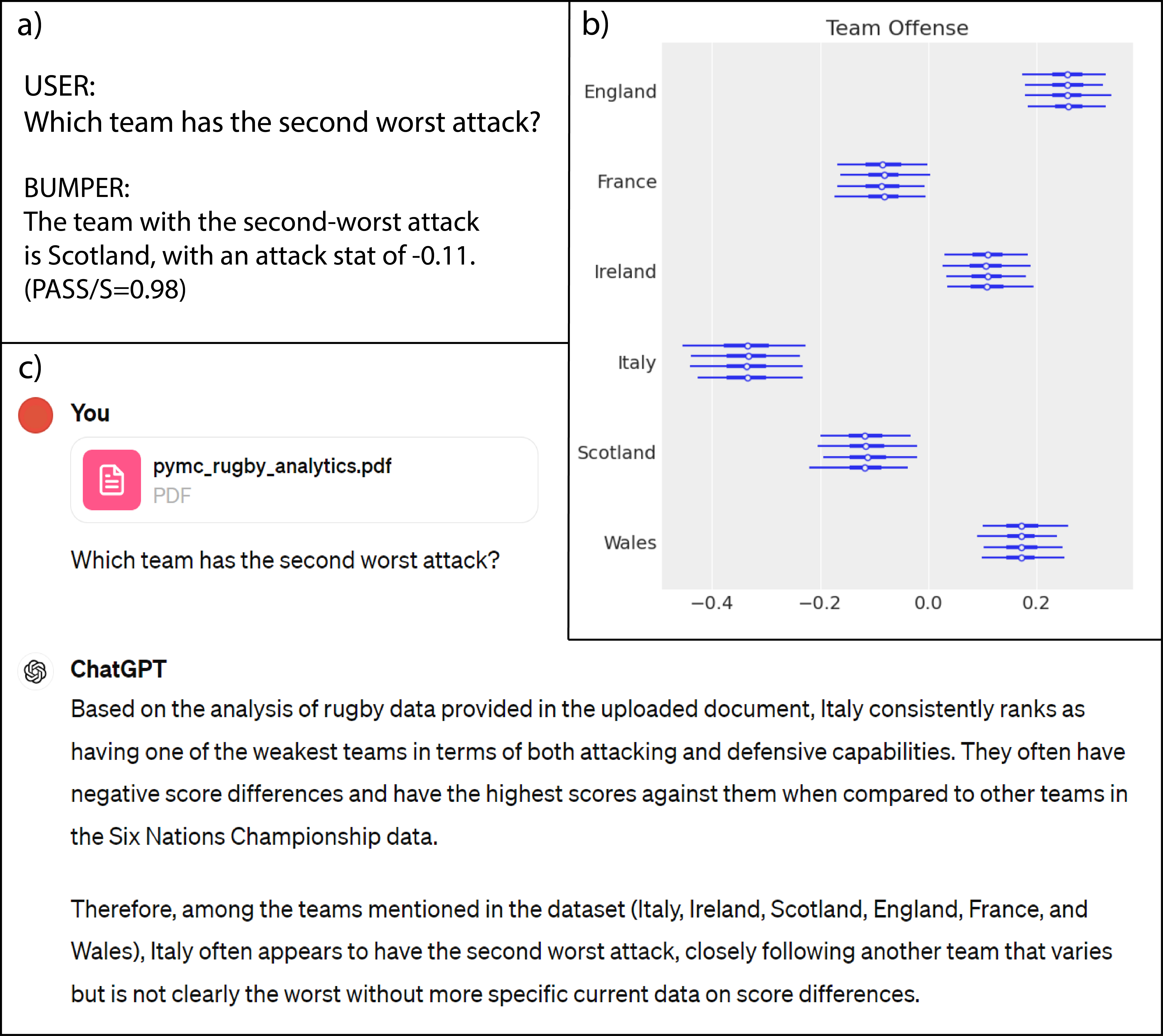}
  \caption{Comparing ChatGPT4 (with context) to a statistical model \NAMEIT. a) Shows the \NAMEIT\,answer. b) contains a figure from the PyMC tutorial which plots the modelled attack statistic for 4 markov-chains \citep{coyle_pymc}. c) Shows answers from ChatGPT4 run with the tutorial as context (run on 5/20/2024).}
  \label{fig:rugby_example}
\end{figure}

\section{Synthesizing evidence: six nations rugby championship example}
\label{sec:rugby}

We use a tutorial from PyMC as a first demonstration and template for interested users (see appendix \ref{appendix:getting_started}). The tutorial describes a Bayesian hierarchical model to analyze and predict outcomes of a rugby tournament (the Six Nations Championship) \citep{coyle_pymc, baio10}. In rugby, two teams play against each other, running with the ball in hand. An ``offense'' or ``attack'' by one team against their opponent is met by a ``defense''. The strength of an offense, in rugby, means how well the team is able attack the opponent.  Historical home and away scores are modeled using a log-linear random effects model with parameters capturing both the attack and defensive strengths of the teams. 

For this \NAMEIT\, example, we constructed two functions to access the estimated attack and defensive strengths of each team, similar to a database query. In Figure \ref{fig:rugby_example}, we demonstrate how this \NAMEIT\,performs against ChatGPT4 with the tutorial as context.
\footnote{https://www.pymc.io/projects/examples/en/latest/case\_studies/rugby\_analytics.html} The user query for this example is: \textit{``Which team has the second worst attack?''}. \NAMEIT\,provides an answer with specificity and clearly related to the provided evidence (e.g., using the estimated model parameters) while ChatGPT4 incorrectly hedges its answer.

\section{Towards trustworthiness: a case study for health policy}
\label{sec:measles}

LLMs are fast evolving into tools that exhibit many of the human-like characteristics that encourage trust and engagement: memory, speech, and multi-modal capability \citep{openai_memory, openai_gpt4o}. However, while these models may affect trust, they have not earned it. In their study establishing trustworthiness benchmarks, Sun et al. \citep{sun24} compare measures of trustworthiness to measures of utility. We contend that in high-stakes situations, trustworthiness is essential for utility.  If you cannot trust a tool, how can you rely on it for critical tasks? We will demonstrate how \NAMEIT\,designs for trustworthiness by applying our framework to an existing, open-source model that estimates the seasonal patterns and outbreak potential for measles \citep{thakkar24b}.

Measles is a highly contagious disease that has a safe and highly effective vaccine \citep{plotkins_measles}.  Currently the World Health Organization (WHO) recommends that every child receive a dose of the measles containing vaccine (MCV) as part of a routine childhood vaccination program but there are many places where the fraction of children who receive this vaccine fall far short of WHO targets (95\%) \citep{wuenic}. In some places, particularly where measles is a constant threat, countries chose to run time-limited, age-targeted supplementary immunization activities (SIAs), or vaccination campaigns. These activities, in conjunction with routine childhood vaccinations, have proven to be an essential tool in avoiding large, catastrophic measles outbreaks \citep{who16, thakkar24a}.

\begin{figure}[ht]
  \centering
    \includegraphics[width=.85\linewidth]{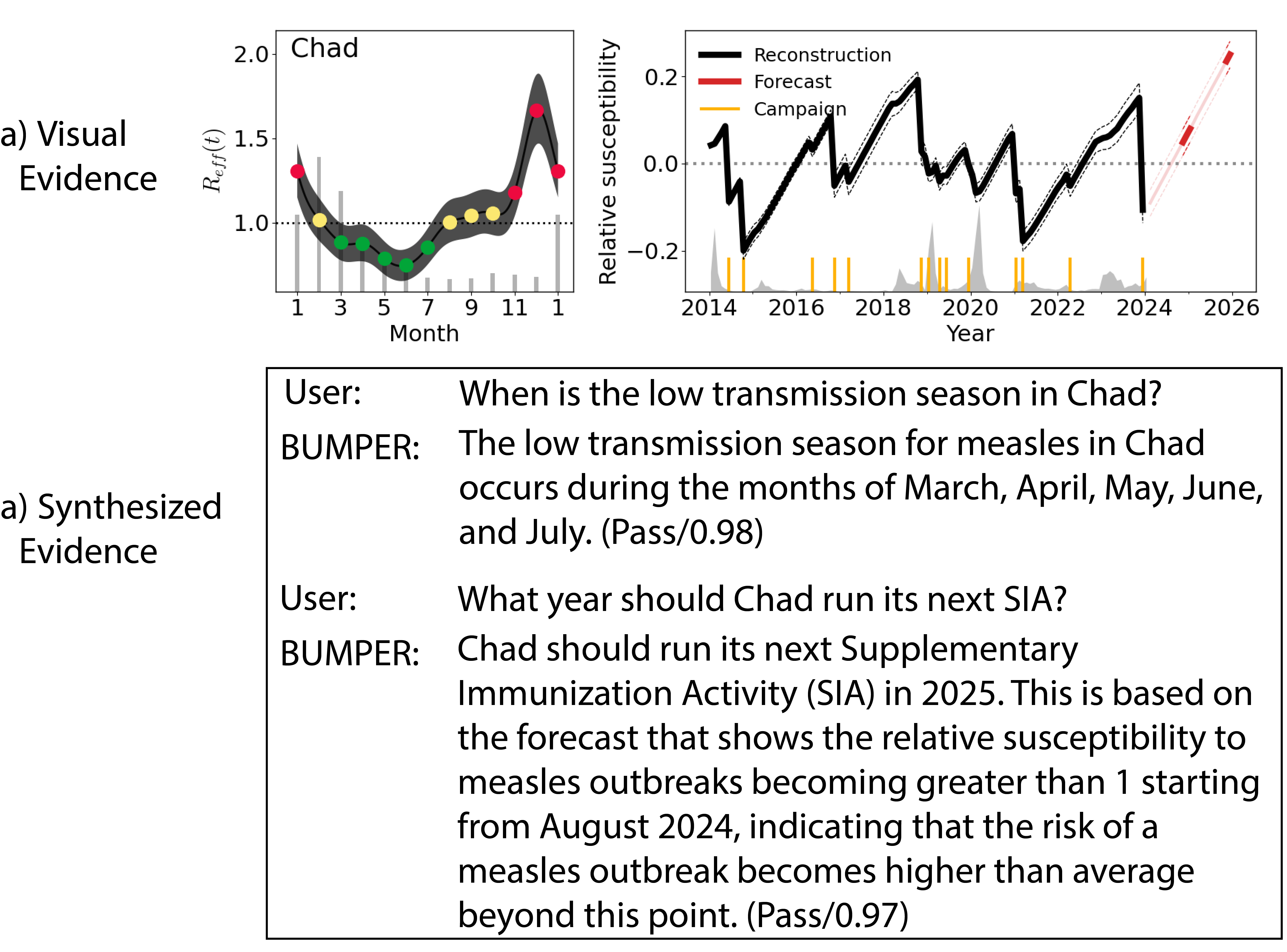} 
\caption{Comparison of visual evidence to textual evidence from BUMPER.}\label{fig:measles_example}
\end{figure}

\begin{figure}[h]
  \centering
  \includegraphics[width=0.9\textwidth]{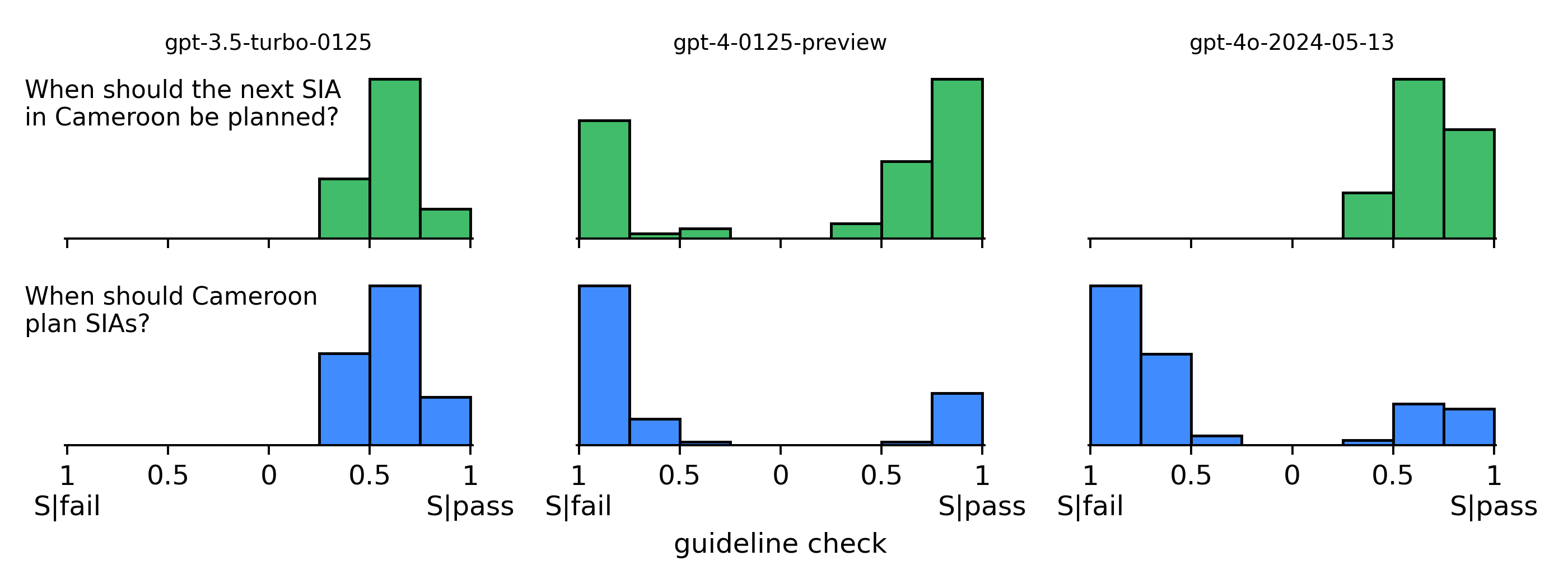}
  \caption{Distributions of compliance scores/token probabilities ($S=P_0$) and overall result (no/yes) returned by the guidelines check. The check is computed against the entire set $G = (c_0, \ldots, c_N) \cup (t_0, \ldots, t_m)$ with no prompt for explanation (see section \ref{sec:framework} details and appendix \ref{appendix:prompts} for examples). The check is called multiple times ($N=3$) for each synthesized answer ($N=25$) for a fixed query ($N=2$; row).}
  \label{fig:singlequery_probs}
\end{figure}

Planning and executing an SIA is challenging and requires coordination at multiple levels of government, attainment of funds from various sources, and management of resources and people. Government officials are key decision makers. Evidence in the form of monitoring, modelling, and and guidelines can help ensure that these campaigns are effective at stopping devastating measles outbreaks\citep{verguet15, who16}. One important consideration is the timing of, or when, an SIA should occur. Timing can be roughly broken down into what year, dictated by the build up of susceptible individuals, and what month, driven by the well-known seasonality of transmission. Plans are determined by individual countries informed by detailed transmission models \citep{thakkar19, zimmermann19} or heuristics \citep{verguet15, who16}. The analysis we chose for this example uses historical measles cases to estimate country-specific seasonality and relative susceptibility (see \ref{fig:measles_example}).

The example's body of evidence is composed of executable code and a paper (see table \ref{tab:bumper_meas}in appendix \ref{appendix:actions}). The code is written in python and fast enough to real in real-time. We wrote wrapper functions around python scripts available on GitHub\footnote{https://github.com/NThakkar-IDM/seasonality} to generate country-specific estimates of: 1) low transmission months, 2) high transmission months, 3) months to run SIAs, and 4) relative susceptibility. For questions regarding methodology we create a separate assistant (see Figure \ref{fig:workflow}) that handles the vector-database query. This allows additional prompt engineering and document targeting not otherwise available through the OpenAI API. Figure \ref{fig:measles_example} provides an example comparing the synthesized, textual evidence provided by \NAMEIT\,with visual evidence that might be found in supplementary material or a dashboard.

\begin{figure}[h]
  \centering
  \includegraphics[width=0.9\textwidth]{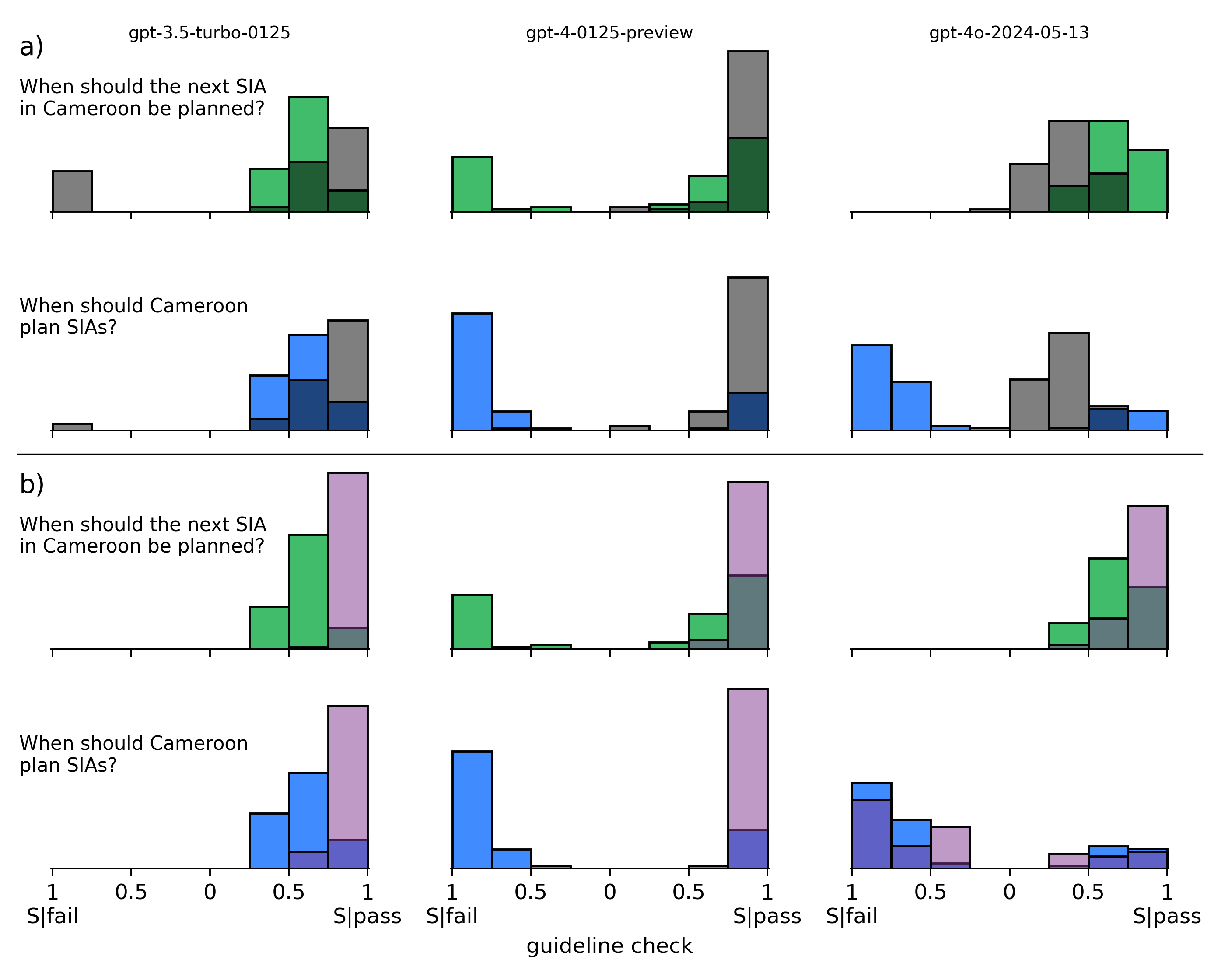}
  \caption{Distributions of compliance scores and overall result (no/yes) returned by the guidelines check.  The check is called is multiple times ($N=3$) for multiple synthesized answers ($N=25$) for a fixed query ($N=2$; row). Blue/green bars are the same as plotted in Figure \ref{fig:singlequery_probs}.a) shows with both the explanation augmented prompt and individual assessments of each guideline element.  b) shows in gray/purple the scores from an explanation augmented prompt. }  
  \label{fig:singlequery_explore_probs}
\end{figure}

The guidelines are listed in appendix \ref{appendix:guidelines}. \NAMEIT\, uses an LLM to determine whether the synthesized evidence is consistent with this statement of purpose (see Figure \ref{fig:workflow}) and records the token log probability as an indicator of confidence \citep{guo17, jiang21}. In Figure \ref{fig:singlequery_probs}a we show the distribution of the token level probabilities ($P_0$) for synthesized evidence answering a single query: a) ``When should the next SIA in Cameroon be planned?'' and b) ``When should Cameroon run SIAs?''.  From inspection of the answers we know that all 3 models do well in answering the question, but this critical step to check whether the answer passes the check see a consistent answer from \gptthree\, but clear bi-modality from \gptfour\, and \gptfouro. Furthermore, comparing across the similarly worded questions we see that the token-level probabilities distributions change. In particular, \gptfouro\,passes answers more consistently than the former. These distributions reveal inconsistencies that undermine \NAMEIT's reliability and must be addressed to ensure its trustworthiness."

Next we prompt the LLM to provide an explanation. The resulting $S=P_0$ distributions are shown in Figure \ref{fig:singlequery_explore_probs}b. Figure \ref{fig:singlequery_explore_probs}a runs an assessment on each element of the guidelines: $c_i$ and $t_i$. The token probabilities are combined as described in section \ref{sec:framework}. We observe that this produces significant change in the $S$ distributions: adding the explanation adds stability to the assessment and checking the guidelines by element reduces the compliance score and increases the spread of $S$.

\begin{figure}[h]
  \centering
  \includegraphics[width=0.9\textwidth]{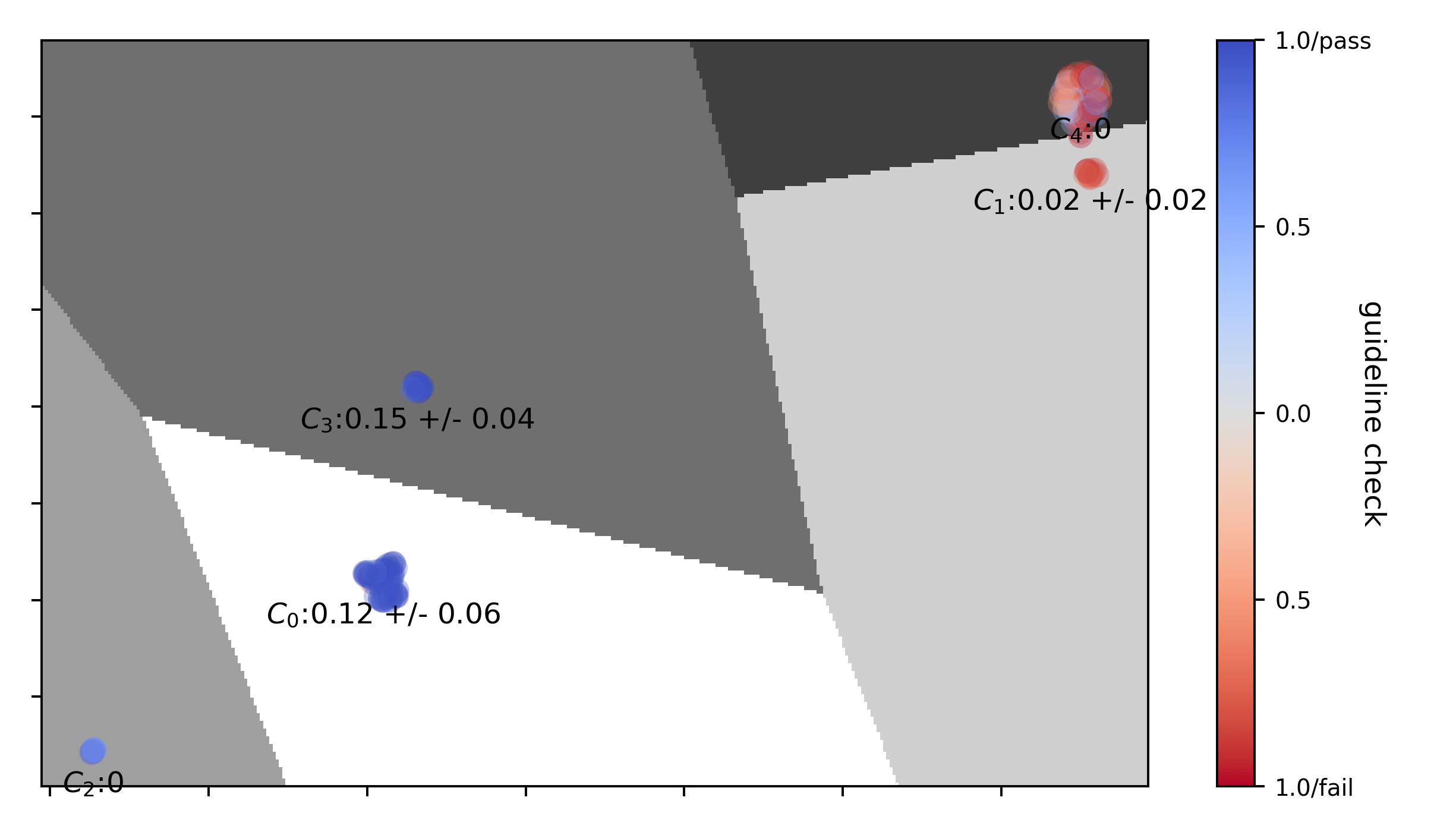}
  \caption{K-means clustering of a UMAP embedding for synthesized answers (\gptfour) to the question: ``When should Cameroon plan SIAs?''. The guideline check was run against the whole guideline and without a prompt for explanation. We have added Jaccard similarity metrics run for 20 samples of the identified cluster for clusters with variation in their answer. This visualization reveals failure modes that appear without the additional prompt and evaluation structure.}
  \label{fig:umap_viz}
\end{figure}

While the compliance score distributions are a helpful indicator of stability, they do not provide much insight into what is causing the various behaviors. Dimensionality reduction techniques combined with clustering can be a powerful tool for developing such intuition. We demonstrate this in figure \ref{fig:umap_viz} where we show a UMAP embedding \citep[2 components, 18 neighbors;][]{mcinnes20} of N=1100 \gptfour\, answers to the question: ``When should Cameroon plan SIAs?''. Each sample plotted is colored by the compliance score generated from evaluation against the whole guideline, $G$, and without a prompt to supply evidence. The embedding, which we established was stable through visual inspection, reveals distinct clusters that are generally consistent in their compliance score. 

We can use K-means to classify the different answers according to the UMAP embedding (see \ref{sec:appendix_umap_examples} for examples). In this example we immediately learn that some of the clusters are replicate answers while other answers vary. This is indicated by the Jacard similarity score calculated between 20 pairs in the cluster that annotates the Figure \ref{fig:umap_viz}. We also learn that the guidelines check is resulting in pass/fail states for some of these clusters but not exclusively (see cluster $C_4$). This supports an interpretation that the evidence prompt (which resolves these issues) is providing both consistency across and flexibility towards evaluating the synthesized answers.

Lastly, the compliance score is part of a suite of checks. Table \ref{tab:checks_meas} provides some examples on various ways that \NAMEIT\,can signal to the user that an answer may not be good. The identified classes, error, out-of-scope, check flag, and check fail, are all associated with choices the scientist makes when assembling their \NAMEIT.

\begin{table}
  \caption{Classes of Check}
  \label{tab:checks_meas}
  \centering
  \begin{tabular}{lll}
    Type     & Example     & Result \\
    \midrule
    
    Error & 
    \makecell[tl]{When should SIAs\\be run in Antarctica?}  & 
    \texttt{ValueError: No data for Antarctica}     \\
    
    Out-of-scope     & \makecell[tl]{Is it more costly\\to run SIAs in\\France or Uganda?} & \texttt{RuntimeError: No tools found}      \\
    
    Check flag    & \makecell[tl]{When should the next\\SIA be run in Chad?} & \makecell[tl]{For Chad, the recommended months for conducting\\supplementary immunization activities (SIAs) against\\measles are July, August, September, and October.\\Measles transmission is high in January, November,\\and December, and is low from March to July.\\Considering the relative susceptibility forecast, which\\indicates an increase starting from July 2024 and\\continuing to rise over the following months, it would\\be optimal to conduct the next SIA in July 2024,\\aligning both with the recommended SIA months and\\the period before susceptibility increases significantly.\\This timing would help mitigate the risk of an outbreak\\by increasing immunity during a susceptible period.\\(PASS/p=80.7)} \\
    
    Check fail & 
    \makecell[tl]{"Is it easier to run SIAs\\in Afghanistan or\\Pakistan?"} &
    \makecell[tl]{For Supplementary Immunization Activities (SIAs),\\the recommended months differ between Pakistan\\and Afghanistan. In Pakistan, SIAs are recommended\\to occur in July and August (months 7 and 8), whereas\\in Afghanistan, the recommended month for SIAs is\\only August (month 8). This suggests that there might\\be a slight flexibility advantage in Pakistan for scheduling\\SIAs over Afghanistan, given the two-month window in\\Pakistan compared to the single month in Afghanistan.\\However, the ease of actually running SIAs will depend\\on a variety of factors, including logistical capabilities,\\accessibility of regions, and the medical infrastructure in\\place within each country. (FAIL/0.97)}\\
    \bottomrule
  \end{tabular}
\end{table}

\section{Discussion}
\label{sec:discussion}

There exist many methods and tools that can query large, multi-modal, databases \citep{pietsch19,llamaindex} and they continue to innovate on, for example, extending attention beyond fixed context windows \citep{dai19, gu22, munkhdalai24} and improvements upon the RAG schema \citep{glass22, asai23, yan24}. These tools are increasingly capable in their ability to ingest any type of media, demonstrating both the exciting potential for communicating information irrespective of medium. There is a rapidly growing body of work applying such tools to scientific knowledge extraction and synthesis \citep{wang24,susnjak24}. However, these studies usually emphasize how to build tools that perform across a field of research. This work presents the BUMPER framework as a tool to be incorporated into and enhance a single study, with the creators retaining clear ownership, enabling accountability and keeping the scientist “in the loop”.

We introduce a framework for facilitating knowledge transfer through the use of LLMs between scientists and decision-makers called the Building Understandable Messaging for Policy and Evidence Review (BUMPER) framework. \NAMEIT\,has several novel features. First, it establishes clear ownership which enables accountability and a simple method for feedback and improvements. Second, it is limited in it scope due to a limited set of actions (e.g., text search, running code, database access) and post-hoc evaluation. Other various guardrail architectures are well into development, but one distinctive element of our proposed solution is that it builds the checks around the synthesized evidence, rather than around the original prompt or LLM. Additionally, it has the potential to help increase accessibility as a digital product and multi-lingual operation.  Lastly, because \NAMEIT\,does not rely on fine-tuning \citep[c.f.][]{zhang24} it has the potential to scale: with the development of standards and increasing the availability of compute and tools, it would be possible to combing multiple \NAMEITs\ into an engine capable of synthesizing from many independent sources.

A critical challenge associated with \NAMEIT, and any tool that tries to use automated evidence generation, is the potential loss of human to human interaction. Translating science to policy well depends on the context which can often only be understood by establishing and maintaining relationships \citep{connelly21}. Therefore it is critical that these tools, while they may reduce some human-to-human interaction, are not taken to be strict replacements. Instead, there are some clear design choices that could potentially increase the quality and availability of relations between scientists and decision makers. For example, attributable and interactive tools can foster follow-up and engagement \citep[e.g.,][]{dontcheva14}.

We have yet to mention the additional burden of work that \NAMEIT\,  - and any tool like it - presents. There are clear benefits to having LLM-enabled translation tools embedded within scientific projects. This established ownership and helps ensure that the correct scientific knowledge is incorporated into the architecture and guideline development. However, this is additional work not currently captured by standard academic measures of success - ``publish or perish'' \citep{hyland16}. Systems could be developed to reward these tools similar to academic software (e.g., Zenodo, JOSS, JORS). But it is also important to note that scientists, who are serious about working with decision makers, have to do a lot of work - serving on panels, advisory commitees, traveling, building and maintaining relationships - that also is not well-captured by citations or publication counts \citep{sie23,r4d}. If the up-front cost of building a \NAMEIT\,can help with some aspects of that work then, in the long term, it may actually reduce overall labor.

Another challenge BUMPER highlights is validation. The task we approach, due to its specialization and purposeful scope limiting, is not well-suited for assessment against established bench-marking datasets like ScienceQA \citep{lu22}. Therefore, it will be essential to formalize validation techniques for high-stakes, specialized LLM applications. Many areas of research offer exciting avenues of further study such as reinforcement learning from human feedback \citep[RLHF;]{russell16,ziegler20}, benchmarks \citep{sun24}, regular expressions \citep{kuchnik23}, and human evaluation \cite{singhal23}. Here we demonstrated how the compliance score could provide insight into stability and performance.

It is also important to consider how open-access vs proprietary LLMs will impact future research from environmental policy to education. Here, due to computational limitations, we have been limited to a proprietary model and associated API. There are benefits in the ability to access other LLMs (e.g., llama3, gemma, mistral-7B) for validation. Open-source models, in particular, can play a role in ensuring accessibility, scalability, and reproducibility \cite{sathish24}. These benefits are well aligned with global calls for open-source science and its relation to policy \citep[][]{unesco_unesco_2021}.
AI promises to produce a sea-change of evidence generation \citep[e.g.,][]{jumper21, koscher23, siebenmorgen24}but it is crucial that this knowledge is made available to affect policy and inform action. Here AI also has the potential to play a significant role by lowering barriers and accelerating synthesis. However, solutions in this domain must be accessible, trustworthy, and accountable; otherwise, the benefits of this technology will be inequitable and slower to appear. \NAMEIT\,offers a clear and actionable path forward to develop solutions addressing these challenges.

\begin{ack}
The authors thank Niket Thakkar, Mandy Izzo, Guillaume Chabot-Couture, and Jonny Tran for their comments. KR, SJ, KM, and JP are employees of the Bill and Melinda Gates Foundation, however, this study does not necessarily represent the views of the Bill and Melinda Gates Foundation.
\end{ack}

{
\small
\bibliography{bibliography}
}
\bibliographystyle{abbrvnat}


\appendix

\section{Appendix / supplemental material}

\subsection{Getting started}
\label{appendix:getting_started}
We implemented \NAMEIT\,using python and provide the code so that interested individuals can try this framework. We have captured the workflow to run the experiments and figures for this using \texttt{snakemake}. The environment was maintained using \texttt{pixi}. To install and run:
\begin{enumerate}
    \item Install pixi: https://pixi.sh/latest/
    \item Install dependencies: \texttt{pixi install}
    \item Run: \texttt{pixi run start}
\end{enumerate}

Individual scripts are found in the scripts/ directory and can be run e.g., \texttt{pixi run python figure\_3.py}. An OpenAI API key is required. 

\subsection{Example actions} \label{appendix:actions}

\begin{table}[h]
  \caption{\NAMEIT\,components for six nation championship}
  \label{tab:bumper_rugby}
  \centering
  \begin{tabular}{llll}
    \toprule
    Source & Access & Purpose & Description \\
    \midrule
    Code  & execution & defensive statistic  & Estimates from the statistical model \\
    Code  & execution & offensive statistic  & Esimates from the statistical model  \\
    \bottomrule
  \end{tabular}
\end{table}

\begin{table}[h]
  \caption{\NAMEIT\,components for measles health policy example}
  \label{tab:bumper_meas}
  \centering
  \begin{tabular}{llll}
    \toprule
    Source & Access & Purpose & Description \\
    \midrule
    Code  & execution & SIA timing  & Months when SIAs are recommended to occur \\
    Code  & execution & high transmission  & Months when transmission is high  \\
    Code  & execution & low transmission  &  Months when transmission is low  \\        
    Paper & vector-storage & methodology & RAG for methodology, sources, etc.      \\
    \bottomrule
  \end{tabular}
\end{table}

\subsection{Guidelines} \label{appendix:guidelines}
We provide the guidelines used for our examples:
\begin{enumerate}

\item Rugby example (section \ref{sec:rugby}):
\begin{fancytextbox}
    Criteria:
    \begin{itemize}
    \item Do not express favoritism
    \end{itemize}

    Topics:
    \begin{itemize}
    \item Rugby
    \item Attack and defense performance estimates
    \end{itemize}        
\end{fancytextbox}

\item Measles example (section \ref{sec:measles}):
\begin{fancytextbox}
    Criteria:
    \begin{itemize}
    \item Do not say anything about any disease besides measles
    \item Do not include any statements regarding cost or financing
    \item Do not make statements saying whether one country is better than another
    \end{itemize}

    Topics:
    \begin{itemize}
    \item Methods or sources
    \item Seasonality
    \item Susceptibility
    \item Supplementary immunization activity (SIA) timing
    \end{itemize}        
\end{fancytextbox}

\end{enumerate}
\subsection{Prompt templates for guidelines check} \label{appendix:prompts}
We used the following prompt templates combining few-shot learning for result consistency.

\textbf{When evaluating synthesized evidence, $E$, entire guideline set, $G = (c_0,\ldots,c_N) \cup (t_0,\ldots,t_M)$}:

\begin{itemize}
    \item No explanation:    
\begin{fancytextbox}
    Does the statement comply with the rule criteria and topics?\\
    Answer "yes" or "no".
\\
    Criteria:\\
    - Do not talk about toast\\
    Topics:\\
    - Whales\\
\\
    Statement: Belugas are blue. \\
    Answer:yes.\\
    ---------------------------\\
    Does the statement comply with the rule criteria and topics?\\
    Answer "yes" or "no".\\
\\
    \{$G$\}
\\
    Statement: \{$E$\}\\ 
    Answer:
\end{fancytextbox}
    \item With explanation:    
\begin{fancytextbox}
    Does the statement comply with the rule criteria and topics?\\
    Answer "yes" or "no" and then explain why.
\\
    Criteria:\\
    - Do not talk about toast\\
    Topics:\\
    - Whales\\
\\
    Statement: Belugas are blue. \\
    Answer:yes. Belugas are not toast and are whales.\\
    ---------------------------\\
    Does the statement comply with the rule criteria and topics?\\
    Answer "yes" or "no" and then explain why.\\
\\
    \{$G$\}
\\
    Statement: \{$E$\}\\ 
    Answer:
\end{fancytextbox}
\end{itemize}

\textbf{When evaluating an individual criteria $c_i$}:
\begin{itemize}
    \item No explanation:
    \begin{fancytextbox}
        Does the statement comply with the rule: "Do not talk about toast"?\\
        Answer "yes" or "no".\\

        Statement: Belugas are blue.\\        
        Answer:yes.\\
\\
        ---------------------------\\
        Does the statement comply with the rule: "\{$c_i$\}" ?\\
        Answer "yes" or "no".\\
\\
        Statement: \{$E$\}\\                
        Answer:        
    \end{fancytextbox}
    \item With explanation:
    \begin{fancytextbox}
        Does the statement comply with the rule: "Do not talk about toast"?\\
        Answer "yes" or "no" and then explain why.\\

        Statement: Belugas are blue.\\        
        Answer:yes. Belugas are not toast.\\
\\
        ---------------------------\\
        Does the statement comply with the rule: "\{$c_i$\}" ?\\
        Answer "yes" or "no" and then explain why.\\
\\
        Statement: \{$E$\}\\                
        Answer:        
    \end{fancytextbox}    
\end{itemize}

\textbf{When evaluating a topic $t_i$}:
\begin{itemize}
\item No explanation:
\begin{fancytextbox}
Is the statement related to the topic: "Whales"?\\
Answer "yes" or "no".\\

Statement: Belugas are blue.\\        
Answer:yes.\\
\\
---------------------------\\
Is the statement related to the topic: "\{$t_i$\}" ?\\
Answer "yes" or "no". \\
\\
Statement: \{$E$\}\\                
Answer:\\    
\end{fancytextbox}
\item With explanation:
\begin{fancytextbox}
Is the statement related to the topic: "Whales"?\\
Answer "yes" or "no" and then explain.\\

Statement: Belugas are blue.\\        
Answer:yes. Belugas are whales.\\
\\
---------------------------\\
Is the statement related to the topic: "\{$t_i$\}" ?\\
Answer "yes" or "no" and then explain. \\
\\
Statement: \{$E$\}\\                
Answer:\\    
\end{fancytextbox}
\end{itemize}

\subsection{UMAP example answers with labels}
\label{sec:appendix_umap_examples}
\textbf{cluster}:$C_0$
\begin{itemize}
\item ($S=0.99$)| For Cameroon, the recommended months for conducting supplementary immunization activities (SIAs) against measles are July and August. Interestingly, these months partly overlap with the low transmission season for measles, which spans March through July, and do not coincide with the high transmission season that occurs in January, September, October, November, and December. Planning SIAs during the low transmission season can be strategically advantageous for controlling the spread of measles effectively.
\item ($S=0.99$)| Cameroon should plan to conduct Supplementary Immunization Activities (SIAs) during the months of July and August. These months also overlap with the low transmission season for measles, which runs from March to July, making it an opportune time to conduct SIAs and prevent outbreaks. Additionally, these months are outside of the high transmission season, which occurs from January and then again from September to December.
\item ($S=0.92$)| In Cameroon, Supplementary Immunization Activities (SIAs) are recommended to occur in July and August. During these months, measles transmission is in a low season, which includes March through July, making it an advantageous period for implementing SIAs to ensure higher coverage and possibly lower transmission risks. The high transmission months for measles in Cameroon are January, September, October, November, and December. Planning SIAs during the low transmission season helps in interrupting the transmission of measles more effectively.
\item ($S=0.95$)| In Cameroon, the recommended months to plan Supplementary Immunization Activities (SIAs) for measles are July and August. This coincides with the low transmission season for measles, which occurs during the months of March, April, May, June, and July. The high transmission season spans from January, and September to December. Planning SIAs during the low transmission period helps to enhance the impact of vaccinations.
\item ($S=0.98$)| For Cameroon, Supplementary Immunization Activities (SIAs) are recommended to occur in July and August. The high transmission months for measles are January, September, October, November, and December. Conversely, the low transmission months are between March and July. Therefore, planning SIAs during the recommended months of July and August aligns well with the period just following the low transmission season, potentially targeting increased susceptibility before entering into the high transmission months later in the year.
\end{itemize}
\textbf{cluster}:$C_1$
\begin{itemize}
\item ($S=-0.87$)| Cameroon should plan Supplementary Immunization Activities (SIAs) for measles in July and August.
\item ($S=-0.84$)| Cameroon should plan Supplementary Immunization Activities (SIAs) in July and August.
\item ($S=-0.89$)| Cameroon should plan Supplementary Immunization Activities (SIAs) for measles in July and August.
\item ($S=-0.67$)| Cameroon should plan Supplementary Immunization Activities (SIAs) in July and August.
\item ($S=-0.56$)| Cameroon should plan Supplementary Immunization Activities (SIAs) for measles in July and August.
\end{itemize}
\textbf{cluster}:$C_2$
\begin{itemize}
\item ($S=0.45$)| Cameroon should plan Supplementary Immunization Activities (SIAs) for measles in the months of July and August.
\item ($S=0.53$)| Cameroon should plan Supplementary Immunization Activities (SIAs) for measles in the months of July and August.
\item ($S=0.85$)| Cameroon should plan Supplementary Immunization Activities (SIAs) for measles in the months of July and August.
\item ($S=0.67$)| Cameroon should plan Supplementary Immunization Activities (SIAs) for measles in the months of July and August.
\item ($S=0.61$)| Cameroon should plan Supplementary Immunization Activities (SIAs) for measles in the months of July and August.
\end{itemize}
\textbf{cluster}:$C_3$
\begin{itemize}
\item ($S=0.98$)| For Cameroon, Supplementary Immunization Activities (SIAs) are recommended to occur in July and August. During this period, measles transmission is historically low (in March, April, May, June, and July), which can facilitate the execution of SIAs by minimizing the risk of disease spread during the campaigns.

The high transmission months for measles in Cameroon are January, September, October, November, and December. Planning SIAs outside these months might help in avoiding peak transmission times, which aligns with the recommended July and August timeframe.

Forecast data on relative susceptibility, a proxy for the risk of measles outbreaks, suggests fluctuating levels through the upcoming months and years. However, the forecast indicates the susceptibility is below 1 until around November 2024, suggesting a lower-than-average risk of measles outbreaks leading up to this point. Beyond November 2024, susceptibility gradually increases, suggesting the risk of outbreaks may become higher than average, reinforcing the importance of conducting SIAs before this projected increase in susceptibility.

It is crucial to plan SIAs during periods that align with low transmission times and take into account future susceptibility forecasts to mitigate the risk of measles outbreaks effectively.
\item ($S=0.97$)| For Cameroon, Supplementary Immunization Activities (SIAs) are recommended to occur in July and August. This is based on available data and aims to achieve optimal impact by timing these activities strategically. 

Looking at measles transmission patterns in Cameroon, the high transmission months are January, and September through December. This period presents a higher risk for measles spread. Conversely, the low transmission months, which are considered a more optimal time for conducting SIAs due to lower measles activity, are March through July.

The forecast for relative susceptibility, which serves as a proxy for the risk of measles outbreaks, indicates that the relative susceptibility is currently below 1 but is predicted to gradually increase over time. By late 2024 and into 2025, the susceptibility index increases, suggesting a higher-than-average risk of measles outbreaks in these future periods. This underlines the importance of planning SIAs accordingly to preemptively tackle this anticipated increase in measles outbreak risk.

Given this information, planning SIAs for July and August aligns with the low transmission season, potentially increasing the effectiveness of these campaigns. Additionally, considering the rising susceptibility forecast, conducting SIAs in these months might also contribute to mitigating the impact of the increasing outbreak risk in the forthcoming periods.
\item ($S=0.98$)| Cameroon should plan Supplementary Immunization Activities (SIAs) in July and August. During these months, measles transmission is generally low, which aligns with the recommendation for conducting SIAs.

The measles transmission in Cameroon is high during January, September, October, November, and December. The low transmission months are March through July, which align perfectly with the recommended SIA months of July and August.

The future relative susceptibility forecast indicates that the risk of measles outbreaks increases gradually from February 2024 onwards, peaking towards the end of 2025. This forecast suggests that planning SIAs before this upward trend in susceptibility becomes significant could be beneficial in mitigating the risk of outbreaks.

Considering the susceptibility forecast and the seasonal pattern of measles transmission, conducting SIAs in July and August would strategically precede the upcoming high transmission season and increased susceptibility, potentially reducing measles transmission and outbreak risk.
\item ($S=0.99$)| For Cameroon, Supplementary Immunization Activities (SIAs) are recommended to occur in July and August. Analyzing the measles transmission patterns and future susceptibility to outbreaks can inform the best timing for these activities:

- **High transmission months** are January, September, October, November, and December.
- **Low transmission months** include March, April, May, June, and July, aligning partly with the recommended SIA months of July and August.

The susceptibility forecast indicates that the risk of a measles outbreak is comparatively lower until approximately November 2024, after which the relative susceptibility increases, indicating a higher than average risk of measles outbreaks. 

Given these factors:
- Conducting SIAs in July and August leverages the period of lower measles transmission, which could help in reaching more children before the susceptibility—and potentially the transmission—increases.
- Planning SIAs before the forecasted increase in susceptibility could also mitigate the risk of larger outbreaks, particularly before the high transmission season resumes in September.

Thus, Cameroon should prioritize planning and executing SIAs during July and August, aligning with the recommendations and considering the forecasted susceptibility to measles outbreaks to ensure efficacy.
\item ($S=0.99$)| For Cameroon, Supplementary Immunization Activities (SIAs) for measles are recommended to occur in July and August. This planning aligns with the low transmission season for measles in Cameroon, which occurs in March, April, May, June, and July, potentially reducing the spread during SIAs.

The measles transmission is high from January and then again from September through December. The susceptibility forecast indicates a relative susceptibility below 1 until November 2024, suggesting the risk of a measles outbreak is lower than average until this time. However, starting from November 2024, the relative susceptibility begins to increase, suggesting that post-November 2024, the risk of measles outbreaks becomes higher than average. 

Given this information, conducting SIAs in July and August is strategically aligned with both the recommended months and a period of lower measles transmission, while also anticipating an increase in outbreak risk later in the year.
\end{itemize}
\textbf{cluster}:$C_4$
\begin{itemize}
\item ($S=-0.54$)| Cameroon should plan Supplementary Immunization Activities (SIAs) for measles in July and August.
\item ($S=0.63$)| Cameroon should plan Supplementary Immunization Activities (SIAs) for measles in July and August.
\item ($S=-0.82$)| Cameroon should plan Supplementary Immunization Activities (SIAs) for measles in July and August.
\item ($S=0.36$)| Cameroon should plan Supplementary Immunization Activities (SIAs) for measles in July and August.
\item ($S=-0.37$)| Cameroon should plan Supplementary Immunization Activities (SIAs) for measles in July and August.
\end{itemize}






\end{document}